\documentclass{article}

\usepackage{amsmath}
\usepackage{amssymb}
\usepackage{xcolor}
\usepackage{booktabs}
\usepackage{multirow}
\usepackage{graphicx}
\usepackage{ulem}
\usepackage{colortbl}
\usepackage{enumitem}
    \PassOptionsToPackage{numbers, compress}{natbib}

    \usepackage[preprint]{neurips_2025}

\usepackage[utf8]{inputenc} 
\usepackage[T1]{fontenc}    
\usepackage{hyperref}       
\usepackage{url}            
\usepackage{booktabs} 
\usepackage{array}
\usepackage{amsfonts}       
\usepackage{amsmath}        
\usepackage{graphicx}
\usepackage{nicefrac}       
\usepackage{microtype}      
\usepackage[skins]{tcolorbox}
\usepackage{wrapfig}
\usepackage{subfigure}
\usepackage{bbm}

\hypersetup{
	colorlinks=true,
	linkcolor=red,
	filecolor=blue,      
	citecolor=blue,
}

\title{APO: Enhancing Reasoning Ability of MLLMs \\via Asymmetric Policy Optimization}

\author{Minjie Hong\thanks{These authors contributed equally to this work.} \quad Zirun Guo\footnotemark[1] \quad Yan Xia\footnotemark[1] \quad Zehan Wang 
\quad \textbf{Ziang Zhang} \quad \textbf{Tao Jin} \quad \textbf{Zhou Zhao}\thanks{Corresponding author.} \\
Zhejiang University \\
\texttt{\{hongminjie,zhaozhou\}@zju.edu.cn}
}

\begin{document}

\maketitle

\begin{abstract}

  Multimodal Large Language Models (MLLMs) are powerful at integrating diverse data, but they often struggle with complex reasoning. While Reinforcement learning (RL) can boost reasoning in LLMs, applying it to MLLMs is tricky. Common issues include a drop in performance on general tasks and the generation of overly detailed or "overthinking" reasoning. Our work investigates how the KL penalty and overthinking affect RL training in MLLMs. We propose \textbf{Asymmetric Policy Optimization (APO)} to address these issues, which divides the sampled responses into positive and negative groups. For positive samples, Difficulty-Adaptive Divergence Shaping (DADS) is introduced to dynamically adjust the KL divergence weight based on their difficulty. This method prevents policy entropy from dropping sharply, improves training stability, utilizes samples better, and preserves the model's existing knowledge. For negative samples, Suboptimal Trajectory Complexity Regularization (STCR) is proposed to penalize overly long responses. This helps mitigate overthinking and encourages more concise reasoning while preserving the model's explorative capacity. We apply our method to Qwen2.5-VL-3B, creating \textbf{View-R1-3B}. View-R1-3B significantly enhances reasoning capabilities, showing an average 7\% gain over the base model and outperforming larger MLLMs (7-11B) on various reasoning benchmarks. Importantly, unlike other reasoning-tuned MLLMs that often degrade on general tasks, View-R1-3B maintains consistent improvement, demonstrating superior generalization. These results highlight the effectiveness and broad applicability of our DADS and STCR techniques for advancing complex multimodal reasoning in MLLMs. The code will be made available at \url{https://github.com/Indolent-Kawhi/View-R1}.
\end{abstract}

\section{Introduction}
Multimodal Large Language Models (MLLMs)~\citep{bai2025qwen25vl, chen2025internvl25, zhu2025internvl3} have demonstrated remarkable capabilities in processing content across language, visual, and audio domains. Despite their impressive performance, these models often struggle with complex reasoning tasks that require nuanced understanding across modalities. Recent advances~\citep{guo2025deepseek, jaech2024openai} have highlighted Reinforcement Learning (RL) as a promising approach to enhance reasoning capabilities in Large Language Models (LLMs), offering a path beyond the limitations of conventional supervised fine-tuning methods such as building a Chain-of-Thought (CoT) dataset~\citep{muennighoff2025s1}. For example, DeepSeek-R1 demonstrates that reinforcement learning with verifiable rewards (RLVR) can substantially enhance mathematical and programming reasoning abilities in language models.

In the multimodal domain, there are some pioneering studies~\citep{peng2025lmm,huang2025vision,meng2025mm} that demonstrate the effectiveness of RL in improving the reasoning capabilities. Despite promising results on the reasoning tasks, these works often exhibit performance degradation on the general multimodal tasks. Additionally, compared to LLMs, MLLMs receive multiple modalities as inputs, which makes it easier for them to generate redundant and repetitive information during the reasoning process. As a result, the overthinking leads to incorrect deduction in the reasoning process, thus providing the incorrect answer finally.

Based on the observations above, we propose an \textbf{Asymmetric Policy Optimization} (APO) that leverages \textbf{Difficulty-Adaptive Divergence Shaping} (DADS) and \textbf{Suboptimal Trajectory Complexity Regularization} (STCR) for targeted optimization of positive and negative samples, respectively, thereby enhancing the reasoning capabilities of MLLMs. We first analyze the KL penalty in the training process and discover that removing it helps the model perform better and learn faster on reasoning tasks, although it degrades performance on general tasks. Conversely, adding the KL penalty retains the generalization capabilities of MLLMs and training stability but sacrifices reasoning abilities. To address this, DADS adaptively adjusts the weight of the correct sample's KL term based on the characteristics of the data samples. This approach not only improves training efficiency and sample utilization but also helps retain the original knowledge of the MLLMs. Additionally, during our experiments, we observe that incorrect answers often involve overthinking. Specifically, we find that the average length of correct responses is consistently longer than that of incorrect ones throughout the training process. Redundant and repetitive responses hinder clear thinking. To address this, we propose the STCR regularization term to penalize incorrect answers with excessively long responses, thereby encouraging the model to generate cleaner and clearer reasoning chains.

We conduct extensive experiments using Qwen2.5-VL-3B~\citep{bai2025qwen25vl} as the base model to develop our reasoning model, View-R1-3B. View-R1-3B demonstrates significant improvements over the base model on various reasoning tasks, achieving an average enhancement of 7\%. Furthermore, it outperforms a range of larger MLLMs, including both closed-source and open-source reasoning models. These results clearly demonstrate the effectiveness of our method in enhancing the reasoning capabilities of MLLMs. Additionally, on multimodal general benchmarks, while other reasoning models experience performance degradation, View-R1-3B maintains consistent improvement, indicating a stronger generalization ability. Our main contributions can be summarized as follows:
\begin{itemize}[leftmargin=*]
    \item We conduct a thorough analysis of the KL penalty and overthinking in the RL training of MLLMs and show that the impact of KL penalty and overthinking on the final performance of MLLMs.
    \item We propose a novel Asymmetric Policy Optimization (APO) consisting of Difficulty-Adaptive Divergence Shaping and the Suboptimal Trajectory Complexity Regularization, which help to improve the training efficiency and sample utilization, and generate cleaner and clearer reasoning chains, respectively.
    \item View-R1-3B outperforms a series of larger MLLMs on both reasoning and general benchmarks, indicating the effectiveness and generalization ability of our approach.
\end{itemize}

\section{Related Work}

\begin{figure}
  \vskip 0.2in
  \begin{center}
  \subfigure[]{\includegraphics[width=.47\columnwidth]{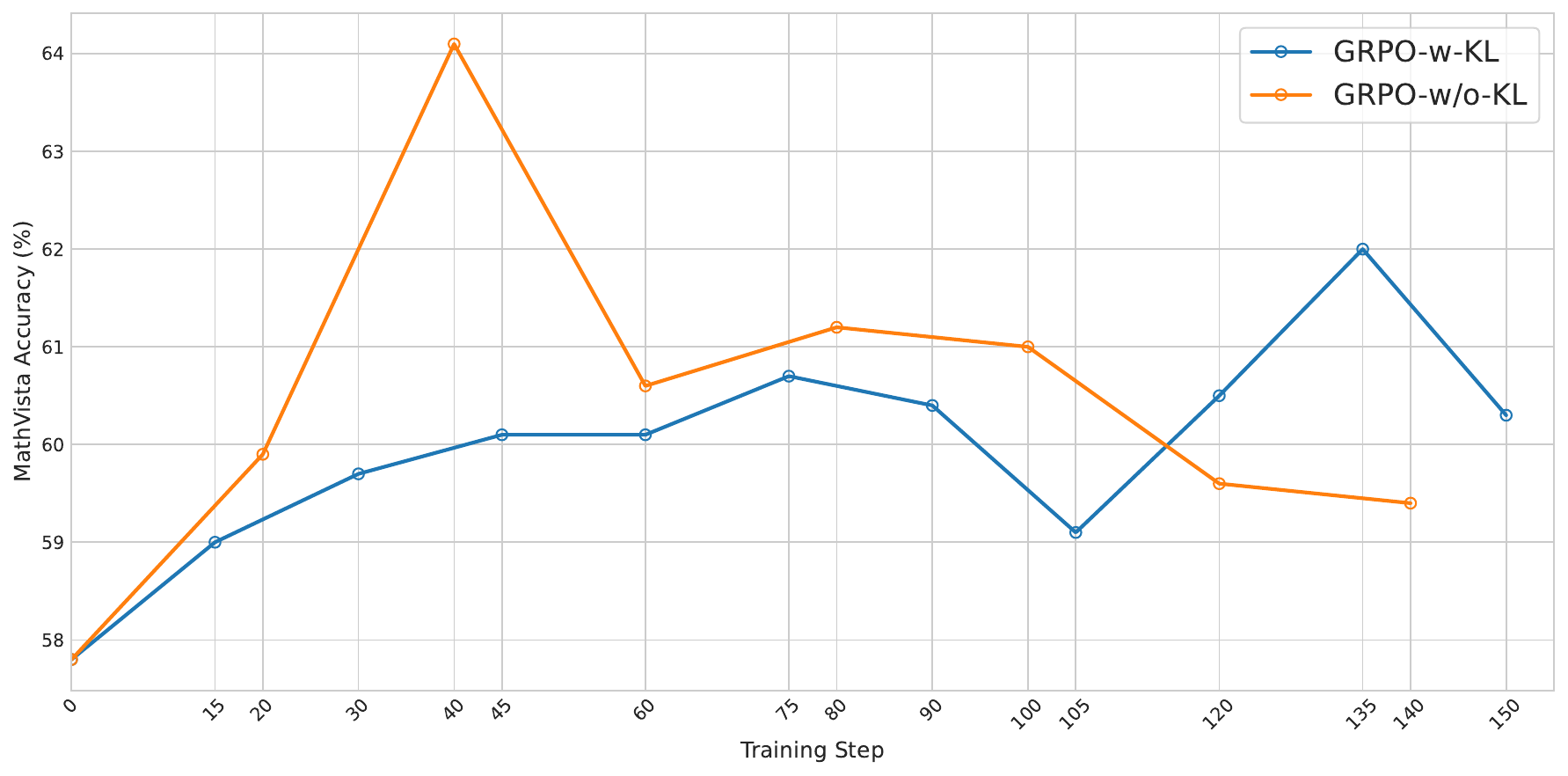}\label{klana:a}}
 \subfigure[]{\includegraphics[width=.51\columnwidth]{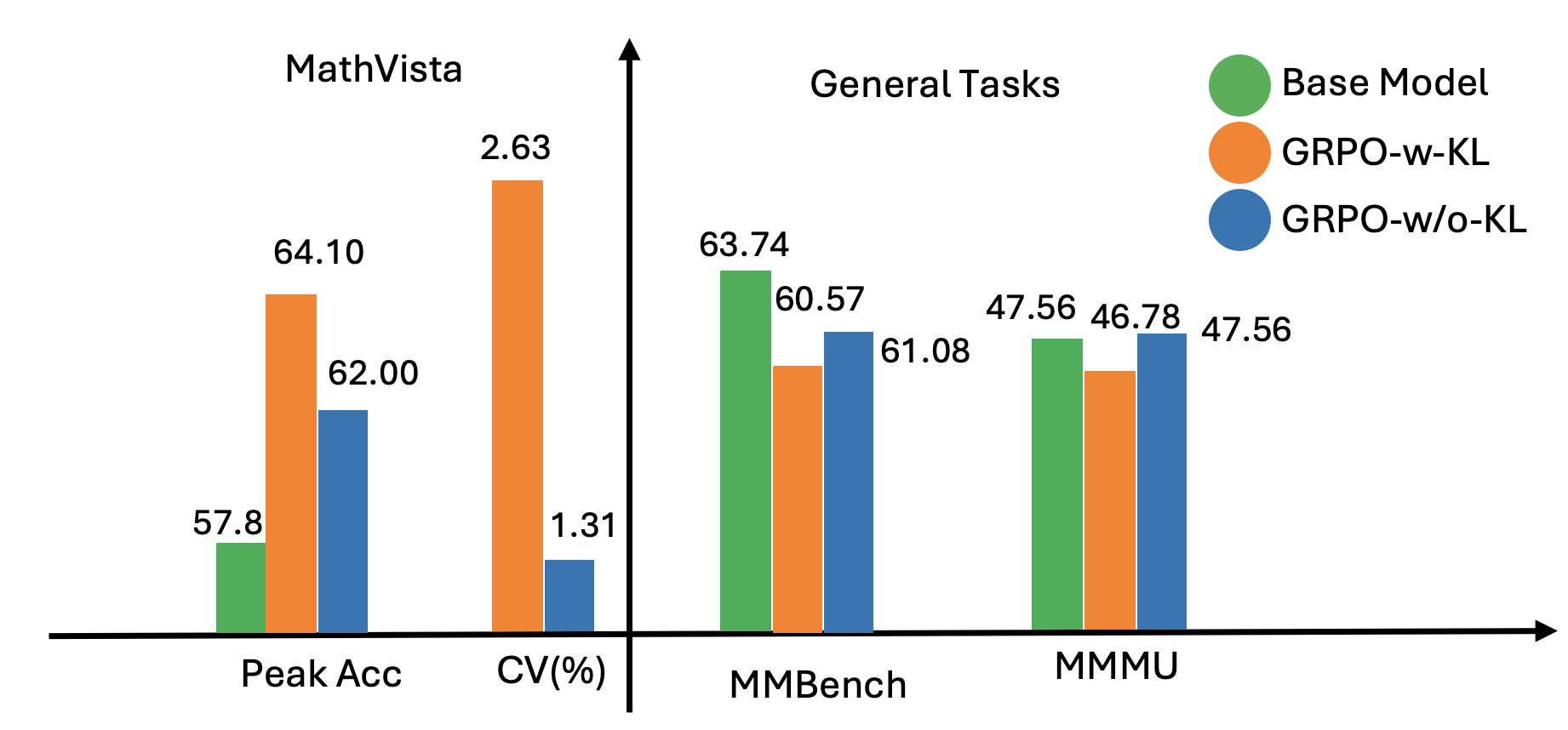}\label{klana:b}\hspace{3pt}}
\\
  \caption{(a) Comparison of the performance on MathVista during the training process. (b) Comparison of the performance and stability on the reasoning and general benchmarks. CV denotes the coefficient of variation of the performance during the training process.}
  \label{klana}
  \end{center}
\end{figure}

\textbf{Reinforcement Learning in LLMs. }
Applying RL to Large Language Models (LLMs) focus from environmental mastery to aligning model outputs with complex, ill-defined human preferences and values. This paradigm, often leveraging Reinforcement Learning from Human Feedback (RLHF)~\citep{ouyang2022training}, fine-tunes pre-trained models by optimizing against human judgments rather than external environment signals.
Proximal Policy Optimization (PPO)~\citep{schulman2017proximal} is an actor-critic RL algorithm that is widely used in the RL fine-tuning stage of LLMs. 
Direct Preference Optimization (DPO)~\citep{rafailov2023direct} eliminates the need for a separate reward model by directly optimizing the language model policy based on a mathematical relationship between human preferences and optimal policies. 
Group Relative Policy Optimization (GRPO)~\citep{shao2024deepseekmath} builds upon PPO by removing the separate value function and evaluating groups of outputs, which is particularly suited for tasks requiring multi-step reasoning and aligning with the comparative nature of reward models. 

\textbf{Reinforcement Learning in MLLMs. }
Extending reinforcement learning paradigms from Large Language Models (LLMs) to Multimodal Large Language Models (MLLMs) has emerged as a prominent research direction. 
Observe-R1~\citep{guo2025observe} enhances MLLM learning through a difficulty-graded dataset, improved visual observation via multimodal constraints, and a reward system for concise, accurate responses with dynamic weighting.
Vision-R1~\citep{huang2025vision} mitigates the issue of "overthinking" through a progressive thinking suppression training stage that progressively expands the length of the Chain-of-Thought (CoT)~\citep{wei2022chain}.
LMM-R1~\citep{peng2025lmm} enhances the model's general task performance by conducting two-stage RL training on mixed datasets of pure text and image-text.
MM-Eureka~\citep{meng2025mm} trains general reasoning in the first stage without imposing a KL divergence constraint, and in the second stage, it incorporates KL divergence to leverage domain-specific data and address performance discrepancies in those domains.
OThink-MR1~\citep{liu2025othink} gradually increases the KL divergence constraint as training epochs progress.
Skywork R1V~\citep{peng2025skywork} employs staged Supervised Fine-Tuning and GRPO~\citep{shao2024deepseekmath} training on a constructed difficulty-graded dataset to achieve adaptive reasoning length based on problem complexity.

\section{Methodology}\label{Methodology}
\subsection{Preliminaries}
\textbf{Reinforcement Learning with Verifiable Reward.}
Group Relative Policy Optimization (GRPO)~\citep{shao2024deepseekmath} is a variant of Proximal Policy Optimization (PPO)~\citep{schulman2017proximal}, which eliminates the value function and estimates the advantage in a group manner. Specifically, given a question-answer pair $(q,a)$, GRPO samples a group $G$ of responses $\{o_i\}_{i=1}^{G}$. Following DeepSeek-R1~\citep{guo2025deepseek}, we use the verifiable reward, which is divided into the accuracy reward $ r_i^{\text{accuracy}}$ and the format reward $r_i^{\text{format}}$. The accuracy reward is the final accuracy of a verifiable task and the format reward is granted when the model follows the format constraint such as \texttt{<think></think><answer></answer>}. Therefore, the total reward $r_i$ of the response $o_i$ can be computed as:
\begin{equation}
    r_i = r_i^{\text{accuracy}} + \lambda r_i^{\text{format}}
\end{equation}
where $\lambda$ is the coefficient to adjust the weight. In GRPO, the reward $r_i$ is normalized in a group manner as follows:
\begin{equation}
    \hat{A}_{i,t}=\frac{r_i-\text{mean}(\{r_i\}_{i=1}^G)}{\operatorname{std}(\{r_i\}_{i=1}^G)}
\end{equation}
Then, GRPO adopts a clipped objective function with the KL penalty term similar to PPO:
\begin{equation}\label{eqgrpo}
\begin{aligned}
    &\mathcal{J}_{\mathrm{GRPO}}(\theta) = \mathbb{E}_{(q,a)\thicksim\mathcal{D},\{o_i\}_{i=1}^G\thicksim\pi_{\theta_{\mathrm{old}}}(\cdot|q)} \\
    &\quad \left[\frac{1}{G}\sum_{i=1}^{G}\frac{1}{|o_{i}|}\sum_{t=1}^{|o_{i}|}\left(\min\left(r_{i,t}(\theta)\textcolor{red}{\hat{A}_{i,t}},\mathrm{clip}\left(r_{i,t}(\theta),1-\varepsilon,1+\varepsilon\right)\textcolor{red}{\hat{A}_{i,t}}\right)-\textcolor{red}{\beta} \mathbb{D}_{\mathrm{KL}}(\pi_{\theta}||\pi_{\mathrm{ref}})\right)\right]
\end{aligned}
\end{equation}
where $r_{i,t}(\theta)$ is the important sampling ratio~\citep{schulman2017proximal}. In this work, we use GRPO as our base approach for RL training. Furthermore, our improvements focus on $\textcolor{red}{\hat{A}_{i,t}}$ and $\textcolor{red}{\beta}$.

\subsection{Difficulty-Adaptive Divergence Shaping}

\textbf{KL or Not?} The KL penalty in Equation~\ref{eqgrpo} is added for the stable parameter update, which can prevent the model from collapsing. However, several recent studies~\citep{meng2025mm, yu2025dapo} propose to remove the KL penalty in the original GRPO for better performance. We pose the question: \textit{ Does removing the KL divergence lead to unintended consequences? } To answer this question and understand the KL penalty in the original GRPO, we conduct several experiments.

Empirically, we present the results with KL penalty and without KL penalty on the reasoning dataset  MathVista~\citep{lu2024mathvista} in Figure~\ref{klana:a}. From the figure, we can observe that the model trained without the KL penalty indeed performs better than the model trained with the KL penalty at most steps. It is easy to understand that without the KL penalty, the model can explore how to reason more freely, breaking away from the constraints of the reference model. This results in higher performance and faster learning in the first several steps. However, as training progresses, the performance of the model with KL penalty starts to decrease while the performance with KL penalty still keeps increasing steadily. This is because, without the KL penalty, the gap between the model and the reference model widens, leading to forgetting previous knowledge, which causes the reasoning ability to gradually degrade. This forgetting can be further validated by the results in Figure~\ref{klana:b}. The figure shows that although the model with KL penalty performs worse on reasoning tasks than the model without KL penalty, it consistently outperforms the latter on general benchmarks. Additionally, we use coefficient of variation, which is calculated as standard deviation divided by sample mean, to measure the stability of the model during the training process. As shown in Figure~\ref{klana:b}, we observe that the performance of the model without KL penalty is not as stable as the that of the model with it.

From the above analysis, we have several key findings. Firstly, removing the KL penalty helps the model perform better and learn faster during the RL training on the reasoning tasks. However, it fails to retain some prior knowledge which leads to performance degradation on the general benchmarks. Secondly, the KL penalty makes the training more stable and helps to improve the reasoning abilities while also retaining the generalization capabilities of the MLLM, but the reasoning ability does not improve as much as that without the KL penalty.

\begin{figure}
    \centering
    \includegraphics[width=1\linewidth]{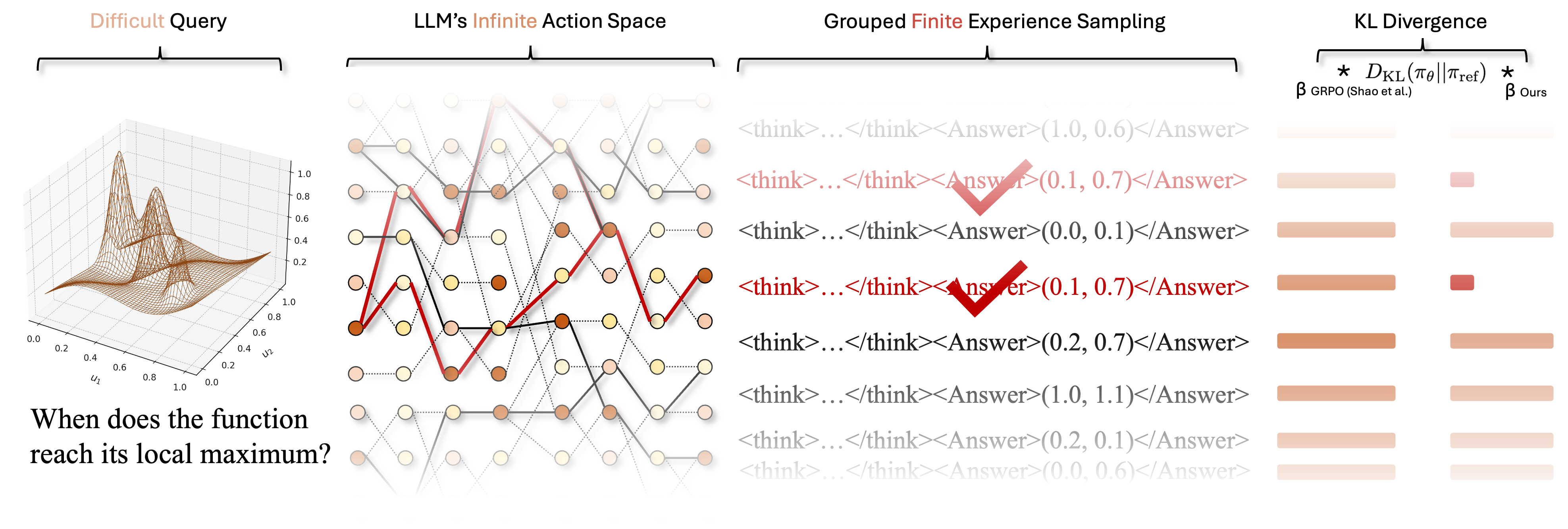}
    \caption{Illustration GRPO and our method dealing with difficult query conditions. DADS adaptively adjusts the KL divergence penalty based on the test-time difficulty of a sample, calculated from a group of sampled responses.}
    \label{fig:KL_main}
\end{figure}

\textbf{Redesign of the KL penalty.} Based on the empirical findings, we need to figure out where to explore and how much to explore, as well as where to retain and how much to retain, in order to achieve the best performance. To this end, we propose the \textbf{Difficulty-Adaptive Divergence Shaping} (DADS) strategy. From Equation~\ref{eqgrpo}, we can observe that for every response, the weighting term $\beta$ has the same value, which means that it imposes equal constraints on all samples. Such a strategy is clearly inappropriate, as different samples and responses have varying training value for improving the reasoning abilities and retaining the prior knowledge. Therefore, DADS is proposed to adaptively shape the KL divergence based on the test-time difficulty of the data sample. Figure~\ref{fig:KL_main} presents the outline of DADS. Specifically, given a question-answer pair $(q,a)$, the current policy samples a group $G$ of responses $\{o_i\}_{i=1}^G$. Then, we define the test-time difficulty as:
\begin{equation}
    d = \frac{\sum_{i=1}^G \mathbbm{1}(r_{i}^{\text{accuracy}} = 0)}{G}
\end{equation}
where $\mathbbm{1}$ is an indicator function evaluated to 1 when $r_{i}^{\text{accuracy}} = 0$, otherwise 0. In GRPO, for very difficult positive samples, even when they yield advantages in subsequent calculations, the imposed KL divergence penalty nonetheless constrains the probabilistic deviation between the actor model and the reference model. This leads to reduced training efficiency for correct samples and makes it prohibitively difficult to master problems that are highly challenging for the frozen reference model. We aim to improve the overall training efficiency of positive samples and to ensure that the KL divergence constraint is weaker for positive samples with higher difficulty, thereby overcoming the limitations imposed by the reference model. Therefore, we propose a projection function $f(d)$ to map the difficulty to the KL weight term adaptively:
\begin{equation}
\beta^{dads}_i =
\begin{cases}
    f(d) \beta & \text{if } r_{i}^{\text{accuracy}} = 1 \text{ and } d \neq 0 \\
    \beta & \text{otherwise}
\end{cases}
\end{equation}

We want $f(d)$ to have several necessary properties: (1) On the interval $[0,1]$, $f(d)$ is a positive-valued function that strictly monotonically decreases from 1 approaching 0. (2) Non-negativity serves as a constraint that helps prevent the actor model from hacking the loss in undesired ways, such as by uniformly decreasing token probabilities. (3) Low initial slopes and terminal fast decays unbind the model under extremely difficult queries. It's worth noting that groups with all-correct samples are filtered out, as decaying the KL divergence of experiences without advantage introduces greater training instability. We select $f(d)=1-\mathrm{e}^{\mathrm{e}(d-1)}$ as the primary function. In addition we explore several variants in Figure~\ref{klall} and analyze them in Section~\ref{Further Analysis}

\subsection{Suboptimal Trajectory Complexity Regularization}

\begin{figure}
    \centering
    \includegraphics[width=1\linewidth]{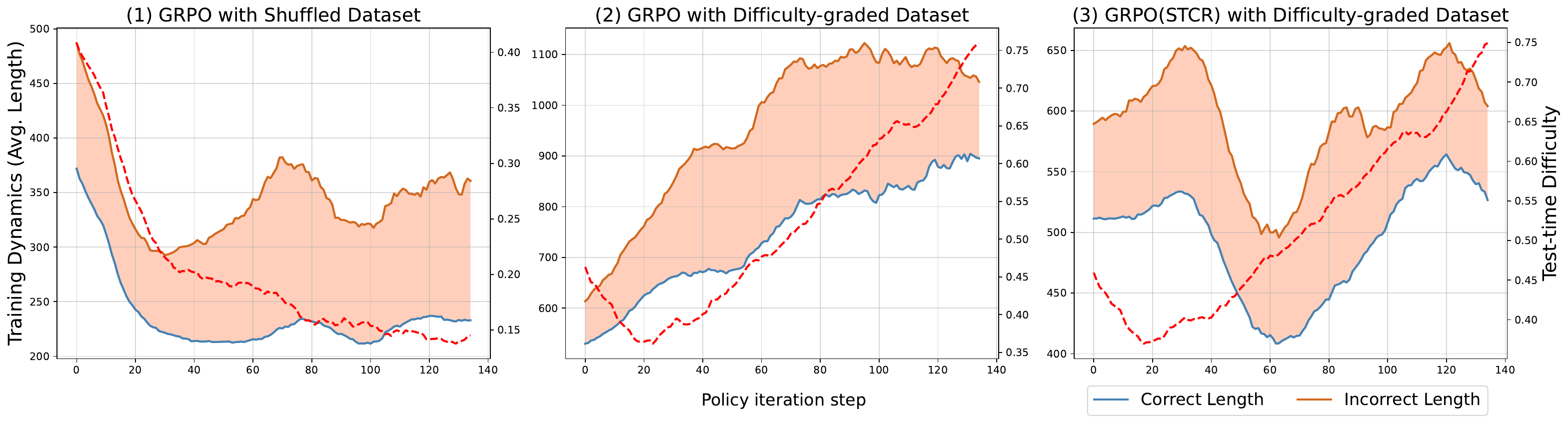}
    \caption{Comparison of response length dynamics during training: Original GRPO on difficulty-graded datasets and shuffled datasets, versus our STCR method applied to difficulty-graded data.}
    \label{fig:length}
\end{figure}

\textbf{Wrong answers always involve overthinking.} In various GRPO-based papers~\citep{shao2024deepseekmath, huang2025vision,meng2025mm,zhou2025r1}, a dramatic increase in response length with training progress is almost universally observed. This is often interpreted as evidence that the model is learning to think or reason more deeply. 
Recently, Dr. GRPO~\citep{liu2025understanding} and DAPO~\citep{yu2025dapo} points out that this may be attributed to the inherent bias of the GRPO objective function. As shown in Figure~\ref{fig:length}, training curves from our GRPO experiments on shuffled dataset and difficulty-graded dataset show that the length of incorrect responses is always longer than that of correct ones, and the difference in length between them continuously increases as the number of training steps increases. We also observe that on the shuffled dataset, the length of correct responses does not increase throughout the training process.

\begin{figure}
    \centering
    \includegraphics[width=1\linewidth]{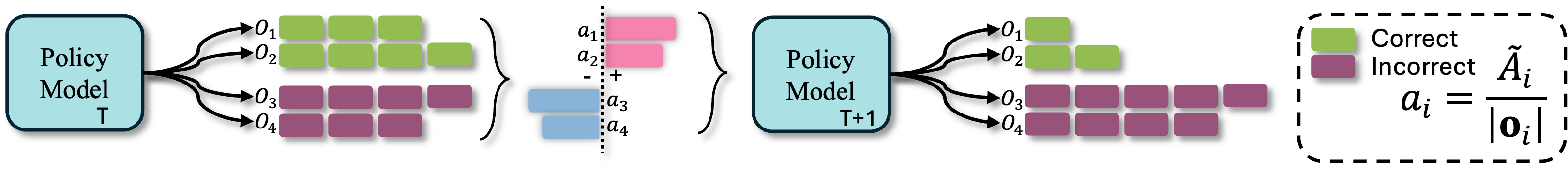}
    \caption{Illustration of the length bias in GRPO. When calculating the effective advantage, the division of $\hat{A}_i$ by the response length $|o_i|$ introduces a length bias to the token-level advantage. This bias preferentially rewards shorter sequences for positive samples and penalizes shorter sequences for negative samples. Over successive iterations, this mechanism can induce a notable divergence in the model's output length, ultimately leading to a phenomenon where incorrect answers progressively lengthen while correct answers become shorter.}
    \label{fig:cmp_main}
\end{figure}

As shown in Figure~\ref{fig:cmp_main}, the original GRPO algorithm assigns equal weight to every sample in the final loss computation. Therefore, when computing the final objective goal, tokens in longer responses have a relatively smaller impact on the overall loss compared to tokens in shorter responses. Over successive iterations, this mechanism can induce a notable divergence in the model’s output length, ultimately leading to a phenomenon where incorrect answers progressively lengthen while correct answers become shorter. Particularly when MLLMs incorporate information from multiple modalities, leading to greater input and output variability, long sequences are more prone to issues like rambling, repetition, and similar problems under response-level length bias~\citep{yu2025dapo}. Therefore, this situation is unacceptable.

Based on these observations, we hypothesize that the length-biased loss function in the original GRPO algorithm contributes to the generation of excessively, particularly for incorrect attempts. This not only wastes computational resources during inference but also dilutes the effective learning signal. To mitigate this issue and encourage the model to generate more concise and accurate responses, we propose a novel regularization technique: Suboptimal Trajectory Complexity Regularization (STCR).

The core idea of STCR is to introduce a penalty term that scales the advantage estimates $\hat{A}_{i,t}$ for suboptimal responses based on their length relative to the average length of correct responses. Specifically, we focus on penalizing incorrect sequences that are significantly longer than the mean length of correct sequences, as these are the primary contributors to the observed length bias and dilute gradient issues.  Let $L_i$ denote the length of the $i_{th}$ sampled trajectory $o_i$ (i.e. $L_i = | o_i|)$, and $L^\mathrm{acc}_\mathrm{mean}$ be the average length of all correctly generated responses within the current batch or training window. For any trajectory $o_i$ where the response is incorrect and its length $L_i$ exceeds $L^\mathrm{acc}_\mathrm{mean}$, we apply a length-dependent scaling coefficient $\alpha_i$ to its advantage estimates $\hat{A}_{i,t}$. The coefficient is computed as:

\begin{equation}
\alpha_i=2-\mu^{L^\mathrm{acc}_\mathrm{mean}-L_i}
\end{equation}

The $\mu$ is a hyperparameter slightly greater than 1. Note that when $L_i > L^\mathrm{acc}_\mathrm{mean}$, the exponent $L^\mathrm{acc}_\mathrm{mean}-L_i$ is a negative value. As $L_i$ increases (meaning the incorrect sequence is much longer than the average correct one), the term $\mu^{L^\mathrm{acc}_\mathrm{mean}-L_i}$ decreases and approaches zero. Consequently, $\alpha_i$ increases towards 2. For incorrect trajectories where $L_i \le L^\mathrm{acc}_\mathrm{mean}$, or for any correct trajectory, $\alpha_i$ is set to 1, applying no scaling. 
This scaling coefficient $\alpha_i$ is then multiplied with the advantage estimate $\hat{A}_{i,t}$ for each token $t$ in the incorrect trajectory $o_i$ only when $L_i > L^\mathrm{acc}_\mathrm{mean}$. The modified advantage term, $\hat{A}_{i,t}^{stcr}$, used in the objective function is:
\begin{equation}
\begin{aligned}
\hat{A}_{i,t}^{stcr} & =
\begin{cases}
\hat{A}_{i,t}\cdot\alpha_i & \mathrm{if~}r^{accuracy}_i=0\mathrm{~and~}L_i>L^\mathrm{acc}_\mathrm{mean} \\
\hat{A}_{i,t} & \mathrm{otherwise}
\end{cases}
\end{aligned}
\end{equation}
By multiplying $\hat{A}_{i,t}$ by $\alpha_i > 1$ for long incorrect sequences, we intensify their negative advantages. This effectively increases the penalty for generating tokens within these overly long, incorrect responses. This targeted scaling amplifies the learning signal, strongly discouraging the model from producing verbose, incorrect outputs.

\begin{figure}
    \centering
    \includegraphics[width=1\linewidth]{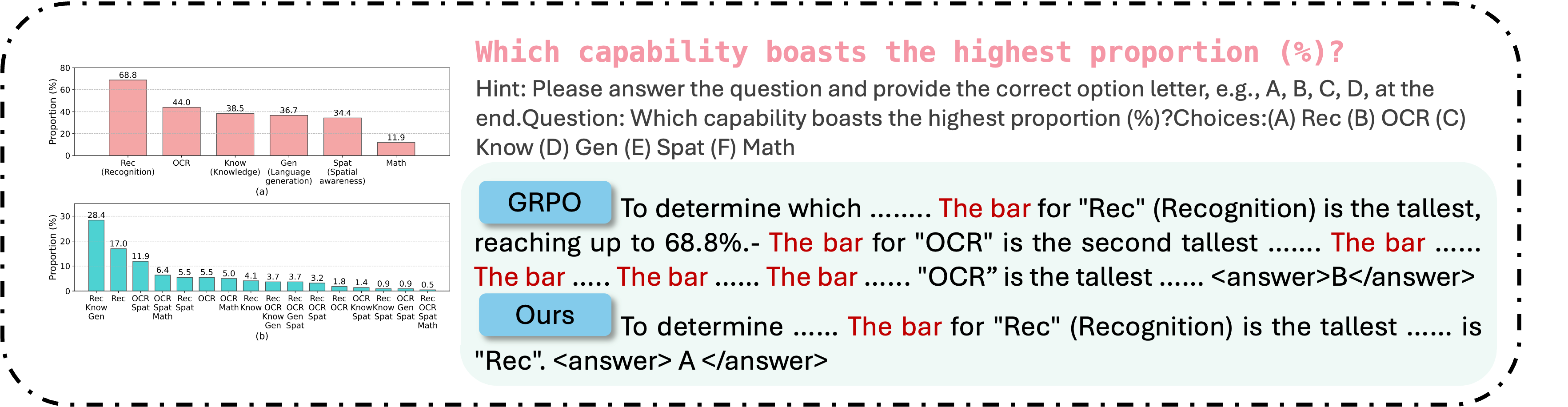}
    \caption{An example demonstrating how our method, by incorporating DADS and STCR, avoids the overthinking seen in the original GRPO, leading to a more concise and correct reasoning process.}
    \label{fig:example}
\end{figure}

\begin{table}[]
\centering
\caption{Comparison with the latest SOTA MLLMs on reasoning and general benchmarks. \textbf{Bold}: best result. \underline{Underline}: second best result.}
\label{mainres}
\resizebox{\textwidth}{!}{%
\begin{tabular}{@{}l|cc|ccc|cc@{}}
\toprule
 &  &  & \multicolumn{3}{c}{Reasoning Benchmark} & \multicolumn{2}{c}{General Benchmark} \\ \cmidrule(l){4-8} 
\multirow{-2}{*}{Model} & \multirow{-2}{*}{Years} & \multirow{-2}{*}{Params} & MathVista & MathVerse & MMK12 & MMStar & MMMU \\ \midrule
\multicolumn{8}{l}{Closed-Source MLLMs} \\ \midrule
GPT-4o~\citep{openai2024gpt4ocard} & 2024 & - & 63.8 & 37.6 & \textbf{49.9} & \textbf{63.9} & \textbf{62.8} \\
GPT-4V~\citep{openai2023GPT4V} & 2024 & - & 49.4 & 39.4 & - & 49.7 & 53.8 \\
GPT-4o-mini & 2024 & - & 52.5 & - & - & 54.8 & \underline{60} \\ \midrule

\multicolumn{8}{l}{Open-Source MLLMs} \\ \midrule
InternVL3-2B~\citep{zhu2025internvl3} & 2025 & 2B & 57.0 & 25.3 & - & 60.7 & 48.6 \\
InternVL2.5-4B~\citep{chen2025internvl25} & 2024 & 4B & 64.1 & 27.7 & - & 58.7 & 51.8 \\
InternVL2.5-8B~\citep{chen2025internvl25}  & 2024 & 8B & 64.4 & 39.5 & 45.6 & \underline{62.8} & 56.2 \\
Qwen2.5-VL-3B~\citep{bai2025qwen25vl}  & 2025 & 3B & 57.8 & 34.7 & 43.5 & 54.1 & 47.6 \\ \midrule
\multicolumn{8}{l}{Open-Source   Reasoning MLLMs} \\ \midrule
LLaVA-CoT-11B~\citep{xu2024llava} & 2024 & 11B & 54.8 & 20.3 & - & - & - \\
Mulberry-7B~\citep{yao2024mulberry} & 2025 & 7B & 63.1 & - & - & 61.3 & 55.0 \\
InternVL2.5-4B-MPO~\citep{chen2025internvl25} & 2024 & 4B & \underline{65.3} & - & - & 58.7 & 51.8 \\
LMM-R1~\citep{peng2025lmm} & 2025 & 3B & 63.2 & 41.5 & - & 58.0 & - \\
R1-VL-2B~\citep{zhang2025r1} & 2025 & 2B & 52.1 & 26.2 & 22.4 & 49.8 & 36.7 \\
R1-VL-7B~\citep{zhang2025r1} & 2025 & 7B & 63.5 & 40.0 & 32.7 & 60.0 & 43.3 \\
R1-Onevision-7B~\citep{yang2025r1} & 2025 & 7B & 64.1 & \textbf{46.4} & 39.8 & 61.9 & 44.4 \\ \midrule
View-R1-3B & 2025 & 3B & \textbf{66.2} & \underline{41.6} & \underline{47.0} & 61.3 & 48.9 \\ \bottomrule
\end{tabular}%
}
\end{table}

\section{Experiments}
\subsection{Experimental Setup}
In our experiments, we use Qwen2.5-VL-3B-Instruct~\citep{bai2025qwen25vl} as our base model. We use multiple datasets to train our model, including MathVision~\citep{wang2024measuring}, We-Math~\citep{qiao2024wemath}, SceMQA~\citep{liang2024scemqa}, PolyMath~\citep{gupta2024polymath}, GeoQA+~\citep{cao2022augmented}, FigureQA~\citep{kahou2018figureqa}, UniGeo~\citep{chen2022unigeo}, TabMWP~\citep{lu2023dynamic}, ScienceQA~\citep{lu2022learn} and CLEVR-Math~\citep{lindström2022clevrmath}. These datasets cover a wide range of domains from reasoning tasks to multimodal general tasks. We filter out all data samples for which the answers could not be verified using our reward function. Then, we use Qwen2.5-VL-3B-Instruct to directly generate the final answer without reasoning for each question and filter out those correct questions which are too simple that might influence the training stability. After filtering, we have around 90k data samples. Due to computational budgets, we randomly sampled 20k of data for our RL training.
To better demonstrate the performance of DADS and SCTR on immediacy datasets with varying levels of difficulty, we construct these datasets with incrementally increasing difficulty according to Observe-R1~\citep{guo2025observe}.

In the training stage, the train batch size is set to 64 and the rollout batch size is set to 128. We use a learning rate of 1e-6 and sample 8 responses for each question. The reward trade-off $\lambda$ is set to 0.5 following~\citep{meng2025mm}. Besides, we use 1.0001 as the hyperparameter value for $\mu$.

For evaluation, we use the reasoning benchmarks MathVista~\citep{lu2024mathvista}, MathVerse~\citep{zhang2024mathverse} and MMK12~\citep{meng2025mm}, and the general multimodal benchmarks MMStar~\citep{chen2024we}, MMMU~\citep{yue2024mmmu}. For MMMU, we use the validation set for evaluation.

\subsection{Main Results}
To fully demonstrate the effectiveness of our method, we compare View-R1-3B with various MLLMs, including closed-source models, open-sourced general models and reasoning models. We present the comparison results in Table~\ref{mainres}. As shown in the table, although View-R1 has only 3B parameters, it achieves the best performance on MathVista and the second best results on MathVerse, MMK12, outperforming a series of closed-source models and 7-11B reasoning models. It fully demonstrates the effectiveness of our method in improving the reasoning capabilities of MLLMs. At the same time, we can observe that most reasoning models perform worse after RL training than the original base model. In contrast, View-R1-3B still exhibits consistent improvements on MMMU, which validates the effectiveness of our method in retaining the original knowledge of MLLMs.

Additionally, we present one example in Figure~\ref{fig:example}. As we can see, the base model lacks a thorough reasoning process and details, which makes the final answer incorrect. The GRPO model is able to generate detailed reasoning process path, but contains much redundant information and outputs, leading to overthinking. As a result,the final answer is incorrect due to the overthinking problem. In contrast, View-R1-3B generates an accurate and concise step-by-step reasoning process.

\subsection{Ablation Study}
\begin{table}[]
\centering
\caption{Ablation study of View-R1-3B, showing the performance on reasoning benchmarks (MathVista and MathVerse) when different components are included. Components evaluated are GRPO, KL penalty, Suboptimal Trajectory Complexity Regularization (STCR), and Difficulty-Adaptive Divergence Shaping (DADS). MathVista sub-scores: Textbook Question Answering (TQA), Arithmetic Reasoning (AR), Statistical Reasoning (SR), Visual Question Answering (VQA). \textbf{Bold}: best result.}
\label{tab:ablation}
\resizebox{0.92\textwidth}{!}{%
\begin{tabular}{@{}ccccc|ccccc|c@{}}
\toprule
\multirow{2}{*}{Model} & \multicolumn{4}{c|}{Variants} & \multicolumn{5}{c|}{MathVista} & \multirow{2}{*}{MathVerse} \\ \cmidrule(lr){2-10}
 & GRPO & KL & STCR & DADS & TQA & AR & SR & VQA & ALL &  \\ \midrule
\multicolumn{1}{c|}{Base Model} &  &  &  &  & 60.1 & 51.0 & 73.8 & 47.5 & 57.8 & 35.0 \\
\multicolumn{1}{c|}{} & \checkmark &  &  &  & 58.9 & 61.8 & 78.1 & 54.7 & 64.1 & 38.9 \\
\multicolumn{1}{c|}{} & \checkmark & \checkmark &  &  & 58.2 & 61.7 & 69.4 & 55.3 & 60.5 & 37.5 \\
\multicolumn{1}{c|}{} & \checkmark &  & \checkmark &  & 58.2 & 62.9 & 78.4 & 53.1 & 64.5 & 40.0 \\
\multicolumn{1}{c|}{} & \checkmark & \checkmark &  & \checkmark & \textbf{65.2} & 60.9 & \textbf{80.1} & 54.2 & 66.1 & 40.1 \\
\multicolumn{1}{c|}{View-R1-3B} & \checkmark & \checkmark & \checkmark & \checkmark & 64.8 & \textbf{63.1} & 78.6 & \textbf{54.9} & \textbf{66.2} & \textbf{41.6} \\ \bottomrule
\end{tabular}%
}
\end{table}

To investigate the contribution of each proposed component to the overall performance of our model, we conduct comprehensive ablation experiments. Starting with the base model, we progressively introduce GRPO, the standard KL penalty, STCR, and DADS. The results are summarized in Table~\ref{tab:ablation}.

Incorporating STCR into the GRPO framework yields notable performance improvements. Specifically, GRPO enhanced with STCR achieves scores of 64.5 on MathVista and 40.0 on MathVerse, surpassing both the basic GRPO and GRPO with the standard KL penalty. This clearly demonstrates STCR's effectiveness in mitigating the length bias inherent in GRPO (Figure~\ref{fig:length}\textcolor{red}{(3)}), promoting the generation of concise and accurate responses, especially for incorrect attempts, thereby enhancing learning efficiency.

Next, we assess the impact of DADS by integrating it into the GRPO framework. This configuration results in a significant performance gain, reaching 66.1 on MathVista and 40.1 on MathVerse. Comparing this result to the performance of GRPO with the standard KL penalty alone underscores the critical role of DADS. These findings support our hypothesis that adaptively adjusting the KL divergence based on sample difficulty enables the model to explore challenging yet correctly answered samples more freely while maintaining stability through stronger constraints on simpler or incorrect samples, as illustrated in Figure~\ref{fig:KL_main}.
Ultimately, our full model, View-R1-3B achieves the highest performance among all configurations, obtaining scores of 66.2 on MathVista and 41.6 on MathVerse. Additionally, it attains the best or near-best results across MathVista sub-benchmarks. The synergy of STCR and DADS effectively addresses the limitations present in standard GRPO, resulting in enhanced performance and improved training dynamics. STCR specifically addresses the length bias, whereas DADS strategically manages the exploration-stability trade-off via adaptive KL divergence shaping.

\subsection{Further Analysis}\label{Further Analysis}
\begin{figure}
  \begin{center}
  \subfigure[]{\includegraphics[width=.31\columnwidth]{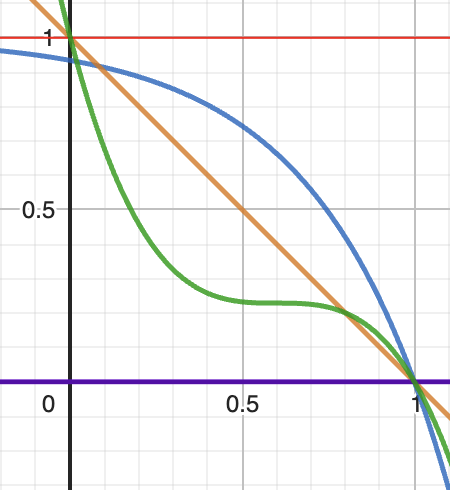}\label{klall:a}}
 \subfigure[]{\includegraphics[width=.335\columnwidth]{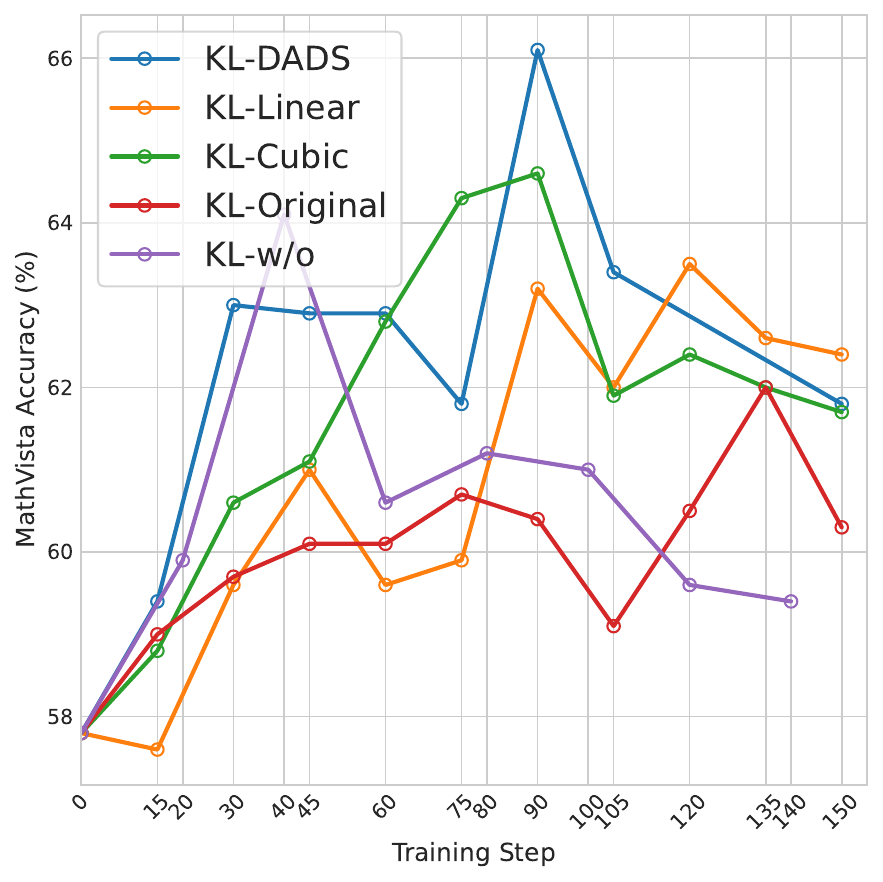}\label{klall:b}}
 \subfigure[]{\includegraphics[width=.335\columnwidth]{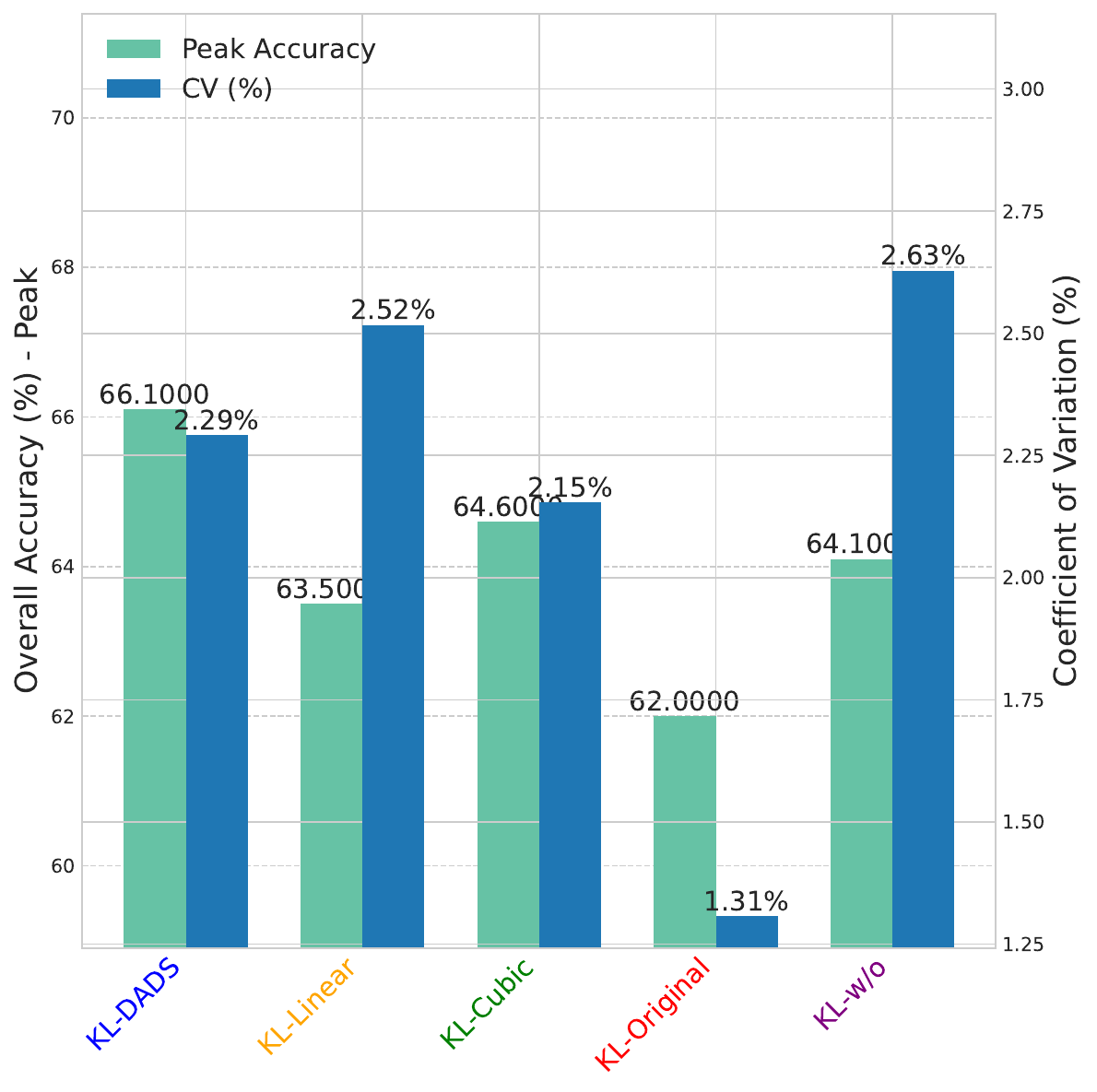}\label{klall:c}}
\\
  \caption{(a) Different curve variants of $f(d)$ in Difficulty-Adaptive Divergence Shaping (DADS). (b) Performance comparison on MathVista during training. (c) Peak performance and training stability (Coefficient of Variation, CV) for different DADS variants. Lower CV indicates higher stability.}
  \label{klall}
  \end{center}
\end{figure}

The core of DADS lies in the projection function $f(d)$, which maps the difficulty $d$ to a scaling factor applied to the original KL weight $\beta$. To validate this and understand the impact of different choices for $f(d)$, we conduct additional experiments comparing linear and cubic variants against the original KL penalty and training without a KL penalty. Figure~\ref{klall} illustrates the performance on MathVista throughout training for these KL penalty strategies. Training without the KL penalty (KL-w/o) leads to rapid initial performance gains but eventually results in performance degradation. In contrast, the original KL penalty (KL-Original) offers stable but slower improvements. Adaptive DADS variants aim to combine these benefits.

We evaluate peak performance (Peak Overall Accuracy on MathVista) and training stability, measured by the Coefficient of Variation (CV) across training steps. Figure~\ref{klall:c} summarizes these results for different configurations. The DADS variants exhibit distinct trade-offs: the linear function $f(d)=1 - d$ enhances peak accuracy (63.5\%) compared to KL-Original but demonstrates lower stability (CV = 2.52\%). The cubic function achieves a superior balance, reaching a higher peak accuracy (64.6\%) than both KL-Original and KL-Linear, while also providing improved stability (CV = 2.15\%) compared to KL-w/o, KL-Linear, and slightly better stability than KL-DADS.

Our proposed KL-DADS variant, employing the function $f(d)=1-\mathrm{e}^{\mathrm{e}(d-1)}$, achieves the highest peak accuracy on MathVista (66.1\%) among all tested variants. Although its stability (CV = 2.29\%) is slightly lower than the original KL and cubic variant, it presents a favorable balance, significantly outperforming training without KL and achieving higher accuracy compared to all other adaptive strategies tested. The function's initial low slope and rapid terminal decay effectively keep penalties high for easier samples, which then quickly drop to zero to encourage exploring harder ones. This strategy ensures sufficient regularization for easier examples, effectively preventing catastrophic forgetting.

We conduct experiments comparing the original GRPO method on both shuffled dataset and difficulty-graded dataset with our STCR method applied to difficulty-graded data, as illustrated in Figure~\ref{fig:length}. The results highlight a consistent issue with the original GRPO: irrespective of dataset difficulty (incremental or shuffled), the length of incorrect responses increasingly diverges from that of correct responses as training progresses. This growing discrepancy visually confirms the overthinking phenomenon, characterized by the model generating excessively long and often repetitive responses when failing to reach correct solutions.

In contrast, when applying our STCR method on the difficulty-graded dataset, Figure~\ref{fig:length} shows that the length difference between incorrect and correct responses remains significantly more stable, staying roughly within a bound of around 100 tokens. This demonstrates that STCR successfully regularizes the length of suboptimal trajectories. Furthermore, as described in the \ref{Methodology}, the hyperparameter $\mu$ in the $\alpha_i$ scaling factor provides a mechanism to control this length difference, allowing for fine-tuning the regularization strength.

The impact of STCR extends beyond just controlling response length; it also positively affects training efficiency and model performance. As shown in the ablation study in Table \ref{tab:ablation}, adding STCR to the base GRPO model improves performance on MathVista ALL from 64.1 to 64.5 and on MathVerse from 38.9 to 40. This improvement indicates that by penalizing overly long, incorrect responses, STCR helps to focus the policy updates on more effective reasoning paths. The regularization encourages the model to generate cleaner and more concise reasoning chains when attempting a problem, which, in turn, can lead to a more efficient learning process and better overall performance. STCR achieves this performance boost while preserving the model's exploration capabilities, unlike approaches that might simply truncate responses or overly constrain the generation process. 

\section{Conclusion}\label{conclusion}
In this paper, we introduce View-R1 and propose a novel Asymmetric Policy Optimization (APO) consisting of Difficulty-Adaptive Divergence Shaping and the Suboptimal Trajectory Complexity Regularization to enhance the reasoning capabilities of MLLMs. Our findings show that adjusting the KL penalty dynamically with DADS improves reasoning without sacrificing generalization. STCR further refines this by penalizing lengthy incorrect responses, promoting clearer reasoning. Using Qwen2.5-VL-3B as a base, View-R1-3B significantly improves performance on reasoning benchmarks, outperforming larger MLLMs while maintaining effectiveness on general multimodal benchmarks. DADS and STCR effectively enhance reasoning while preserving generalization, setting the stage for more capable MLLMs in diverse applications. Due to resource constraints, we cannot currently expand the model to a larger size.

{
\small

\bibliographystyle{plain}
\bibliography{ref}
}


\appendix

\end{document}